\documentclass{svproc}

\usepackage{graphicx}
\usepackage{url}

\usepackage[margin=0.8in]{geometry}

\usepackage{wrapfig}
\usepackage{algorithm}
\usepackage[noend]{algpseudocode}
\algnewcommand\algorithmicforeach{\textbf{for each:}}
\algnewcommand\ForEach{\item[ \algorithmicforeach]}

\begin{document}
	
	\mainmatter
		
		\title{Random Subspace Mixture Models \\for Interpretable Anomaly Detection}		
		\author{Cetin Savkli\inst{1} \and Catherine Schwartz\inst{1,2}}
                      \authorrunning{Cetin Savkli et al.}
                     \tocauthor{Cetin Savkli, Catherine Schwartz}

		\institute{Applied Physics Lab, Johns Hopkins University, Laurel, MD, USA\\	
                                      \email{cetin.savkli@jhuapl.edu}\\
                                  \and  Department of Mathematics, University of Maryland, College Park, MD, USA\\
                               		\email{catherine.schwartz@jhuapl.edu}}
		
\maketitle

		\begin{abstract}
We present a new subspace-based method to construct probabilistic models for high-dimensional data and highlight its use in anomaly detection. The approach is based on a statistical estimation of probability density using densities of random subspaces combined with geometric averaging.  In selecting random subspaces, equal representation of each attribute is used to ensure correct statistical limits.  Gaussian mixture models (GMMs) are used to create the probability densities for each subspace with techniques included to mitigate singularities allowing for the ability to handle both numerical and categorial attributes. The number of components for each GMM is determined automatically through Bayesian information criterion to prevent overfitting. The proposed algorithm attains competitive AUC scores compared with prominent algorithms against benchmark anomaly detection datasets with the added benefits of being simple, scalable, and interpretable.
		\end{abstract}

\section{Introduction}
As availability of data and applications of machine learning continue to grow, the need for simple and scalable algorithms become increasingly necessary. While the practice of machine learning has a long history, the complexity and scale of data along with the human expertise necessary to implement and execute analyses continue to present a challenge. In recent years, the need for interpretable models that provide insights into their decisions has also emerged as an important factor that contributes to the acceptance of analytic results by potential consumers. 

Probabilistic models are one of the most fundamental and elegant tools in the machine learning repertoire, providing a concise representation of data and serving as a basis for common machine learning applications such as classification and anomaly detection. They are particularly suitable for unsupervised anomaly detection, a problem that is increasingly encountered as the labelling of large quantities of data is often not available or feasible. Probabilistic models are intrinsically transparent and their results are explainable as they essentially capture the statistics of the data. 

One of the most well-known approaches to probabilistic modeling are mixture models~\cite{MclachlanBasford88}. Mixture models capture the distribution of data and the complexity of relationships between attributes through a superposition of simple multivariate distributions. They provide a structure that facilitates the measurement of probability density and are also suitable for understanding clustering of data. Mixture models have been successfully used for numerical as well as categorical data~\cite{Savkli2017GALILEOAG} and are particularly suitable for datasets that have moderate to low dimensionality. 

Probabilistic graphical models are another important category of probabilistic models that have gained popularity in recent years. They are particularly suitable for high-dimensional datasets as they factorize the joint probability distribution into attribute subspaces according to an analysis of the dependencies between attributes. Examples of these models are junction/clique trees~\cite{1054142,koller2009probabilistic,savkli2016bayesian} and Bayesian networks~\cite{10.5555/52121}. Subspace-based decomposition, also known as structure learning, underlies probabilistic graphical models and is often based on a correlation metric such as mutual information or Pearson correlation. While probabilistic graphical models reduce the dimensionality of the problem by capturing interdependencies between attributes, the resulting structure is not naturally guaranteed to be a balanced partitioning of the attribute space which would be helpful in facilitating parallelization and scalability. 

As far as high-dimensional data analysis is concerned, subspace-based approaches are particularly appealing as they can potentially reduce the complexity of model building and lead to scalable solutions. Examples of such approaches include outlier detection based on the combination of subspace anomaly scores~\cite{lazarevic2005feature} as well as ensembles of weak learners such as those used in random forest~\cite{Statistics01randomforests}.

The goal of this paper is to present a new and simple probabilistic model, combining desirable features of mixture models and subspace-based methods, that can easily be applied to high-dimensional and large-scale datasets. We seek to create a model that (i) is easily parallelizable, (ii) is simple to implement, (iii) provides results that are easily interpretable, and (iv) can be used for numerical, categorical, and mixed datasets. To accomplish these goals, we present a new model, \textbf{Random Subspace Mixture Model (RSMM)}, which leverages some of the useful features of existing methods. RSMM is based on two key assumptions: 
\begin{itemize}
	\item The joint probability distribution is constructed from a geometric average of probability densities of $m$ randomly selected subspaces of dimension $k$ such that each dimension of the data is represented in an equal number of subspaces in the final probability distribution. 
	\item Each subspace distribution is constructed on a standardized dataset with added noise using a multivariate mixture model through expectation maximization~\cite{Dempster77maximumlikelihood} where the number of mixture components being determined by using an optimality criterion such as AIC~\cite{akaike1998information} or BIC~\cite{schwarz1978estimating}.
\end{itemize}

These features lead to a balanced factorization of the joint probability distribution that facilitates parallelization and does not require a complex step regarding constructing the model structure. Furthermore, the subspace-based structure of the probability distribution contributes to the interpretability of the results. While the spirit of the algorithm is analogous to other subspace methods, our main focus here is to construct a probabilistic subspace-based model for the data that can be used for anomaly detection. 

Anomaly detection refers to the problem of finding patterns in data that do not conform to expected behavior~\cite{chandola2009anomaly}.  It includes both outlier detection and novelty detection and is a field of study that has been around for a long time~\cite{zimek2018there}.  Anomaly detection has been used to find irregularities in many different domains including in hyperspectral imagery~\cite{ettabaa2019anomaly}, maritime surveillance~\cite{riveiro2018maritime}, healthcare analytics~\cite{ukil2016iot}, medical applications~\cite{wei2018anomaly}, and the Internet of Things~\cite{sipple2020interpretable}.  It is also used for fraud detection~\cite{ahmed2016survey}, intrusion detection~\cite{greggio2018anomaly}, flight operation and safety monitoring~\cite{li2016anomaly}, and to detect misinformation~\cite{kumar2019detecting}. The proposed RSMM algorithm is a generic algorithm that can easily be applied to detect anomalies in any domain or application.

\section{Random Subspace Mixture Model}

The RSMM is developed to work with datasets $X_{s \times n}$ where $s$ represents the number of data points in the dataset, also referred to as its size, and $n$ represents the number of attributes in the dataset, also referred to as its dimensionality. 

The base of the proposed algorithm are mixture models. A mixture model is defined by a superposition of probability densities for $c$ components,
\begin{equation}
f(x) = \sum_{i=1}^c f(x \vert C_i) p(C_i),
\end{equation}
where $x=(x_1,x_2,\cdots , x_n)$ and each component distribution, $C_i$, is subject to the normalization condition,
\begin{equation}
1 =  \int d^nx f(x \vert C_i),\label{normalization}
\end{equation}
and the components priors determine the relative size of each components,
\begin{equation}
1 =  \sum_{i=1}^c p(C_i).
\end{equation}
Gaussian mixture models (GMMs) are a special case where individual components are defined as multivariate normal distributions:
\begin{equation}
f(x\vert C_i) \equiv N(x;\Sigma_i,\mu_i)
\end{equation}
where $\Sigma_i$ and $\mu_i$ represent an $n \times n$ covariance matrix and an $n$ dimensional vector that represents the mean of component $i$. 

While the specification of a mixture model is concise and general, when the dimensionality of the data is large a variety of numerical challenges arise. One challenge is associated with numerical instabilities that arise when working with large-scale matrices. As dimensionality of the data grows, sparsity of the distribution can result in pancake-like mixture components causing singularities. The second challenge is related to the difficulty of parallelization. Combining all attributes into a single multivariate model makes parallelization difficult to implement and limits the application of the algorithm to smaller datasets of lower dimensionality. Finally, although the parameters of the distribution and mixture components provide a detailed specification of the model, when inference is performed using a high-dimensional probability density model, gaining insights about what combination of factors explain the result requires additional effort. 

\subsection{Combination of Subspace Probability Densities}
\label{subspace_combo_section}

To address these problems, we follow an ensemble approach to model the joint probability density by using a rescaled geometric averaging of the probability densities of $m$ lower dimensional subspaces of dimension $k$. While subspaces are selected randomly, the geometric averaging has to be done in a way that reduces to known statistical limiting cases such as Naive Bayes limit for $(m,k)=(n,1)$ and full joint probability distribution for $(m,k)=(1,n)$. The rescaled geometric averaging that meets these requirements can be defined as:
\begin{equation}
rsmm(x) \equiv \left( \prod_{i=1}^{m}gmm^{n/k}_i(x)\right)^{1/m}\label{geometric}
\end{equation}

\begin{wrapfigure}{r}{0.5\linewidth}
	\vspace{-10pt}
%\begin{figure}[htbp]
	%\centerline{\includegraphics[width=100mm]{figures/geometric_mean_wide1D_legend_top}}
\includegraphics[width=0.48\textwidth]{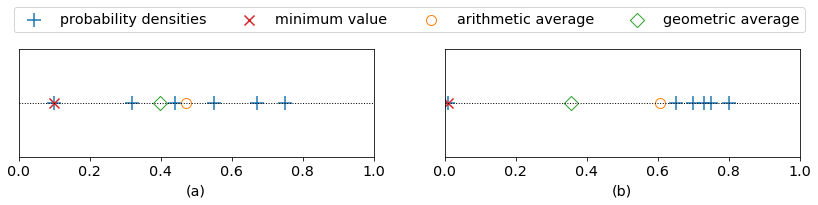}
	\caption{Toy scenarios demonstrating the combination of subspace probability densities that are (a) diverse and (b) clustered with a single near-zero value.}
	\label{fig_geo_ar_min}
%\end{figure}
\end{wrapfigure}

The $n/k$ addition is the only modification to the typical defintion of geometric averaging.  While the $n/k$ factor does not change the ordering of the densities, it guarantees the probability density to reduce to known limiting cases.  Besides having desirable limits, geometric averaging of subspace probability densities, as opposed to arithmetic averaging or using the minimum value, captures anomalies found in different subspaces more effectively. As illustrated with two toy scenarios in Figure~\ref{fig_geo_ar_min} (analogous to Figure 2 in~\cite{zimek2014ensembles}), geometric averaging is able to be heavily influenced by very low probability densities in a few subspaces while still accounting for the probability densities across all subspaces. In practice, to prevent a probability density of 0 from setting the combined result to 0 and prevent ranking of anomalies between the most anomalous events, a machine epsilon is added to every subspace probability density. The expression in Equation~\ref{geometric} can be used for ranking the densities for anomaly detection purposes. 

An ensemble of low-dimensional subspaces where each subspace has the same dimension $k$ also provides a result that is interpretable. When the model labels a point to be anomalous, attributes can be ranked based on how often they are part of subspaces that characterize the point as anomalous compared to how often they are part of subspaces that characterize the point as normal. Additionally, the result can be explained in terms of subspace probabilities, a feature that is similar to probabilistic graphical models.

\subsection{Selection of Subspaces}

The first consideration in selection of subspaces is equal participation of attributes in the subspaces which is necessary to ensure the correct statistical limiting behavior described in the previous section. In order to provide an equal participation for each of the coordinates in the subspace distributions, the specification of the number of subspaces $m$ has to obey the condition:
\begin{equation}
	m \in \alpha \times \frac{LCM(k,n)\label{msub}}{k} \label{setm}
\end{equation}
where $LCM$ represents least common multiple and $\alpha$ is an integer. Constructing $k=6$ dimensional subspace models for an $n=20$ dimensional data, the available $m$ values would be $m \in 10, 20, \cdots$.

Each subspace distribution $gmm_i(x)$ in Equation \ref{geometric} is then built by randomly selecting sets of $k$ distinct attributes for each of the $m$ subspaces, ensuring that no duplicate subspaces are created.  Some of the subspaces, such as those that entirely reside in a categoric subspace, are more elementary than others and addressing this may further simplify the implementation. At this time, we have not taken advantage of this and have treated every subspace equally.  Categorical attributes are discussed more in Sections \ref{mitigations} and \ref{subspace_study}.

\subsection{Determination of Subspace Component Number}

For each subspace, a GMM is obtained with a number of components, $c$, that is determined automatically by the algorithm. For each proposed value of $c$, an EM procedure is performed to converge to a solution $gmm_{i}^c(x)$ using the available components provided. The quality of the mixture model solution can be measured using metrics such as the Akaike Information Criterion (AIC,~\cite{akaike1998information}) and Bayesian Information Criterion (BIC,~\cite{schwarz1978estimating}). These are given by
\begin{equation}
{\rm AIC} = 2\nu - 2\log(gmm_{i}^c(x))
\end{equation}
\begin{equation}
{\rm BIC} = \log(s)\nu-2\log(gmm_{i}^c(x))
\end{equation}
where $\nu$ is the degrees of freedom of the model.
A detailed description and comparison of these metrics is given by~\cite{vrieze2012model,aho2014model}.

In this work, the number of components $c$ of a subspace is determined by searching for a minimum BIC with the lowest component number to prevent overfitting of subspace distributions. The search is performed starting with $c=1$ and incrementing the number of components until the BIC score starts increasing. Since mixture models are dependent on initial conditions, the number of initiations to perform for each GMM is set to 3 to minimize influence of the initial state and to help ensure the minimum BIC is chosen appropriately. The measures taken to prevent overfitting with too many components helps the algorithm handle contaminated data in which the training data contains both normal and anomalous data points. A global optimization that varies all subspace parameters in a coordinated way to minimize a global BIC score that is defined using the joint probability distribution is computationally not feasible and has not been attempted. It is however possible to vary number of subspaces as well as subspace dimensionality through a grid search where these 2 parameters are automatically determined.

\begin{wrapfigure}{r}{0.5\linewidth}
\vspace{-25pt}
%\begin{figure}[thb]
	%\centerline{\includegraphics[width=0.5\columnwidth]{figures/componentFrequency.png}}
\includegraphics[width=0.5\columnwidth]{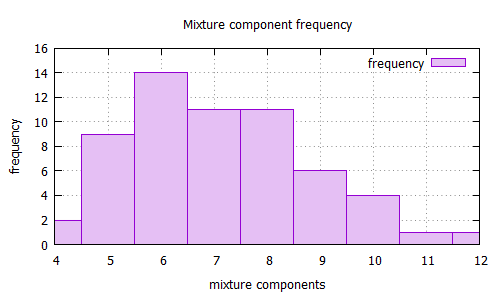}
	\caption{This figure illustrates the number of mixture components and their frequencies encountered in building an RSMM model using the well known Mushroom dataset. Each susbpace provides an alternative partitioning of the data according to its own mixture components and further analysis can be performed to cluster data according to subspace results.}
	\label{fig_component_freq}
%\end{figure}
\end{wrapfigure}

An additional benefit of the RSMM model is in facilitating alternative clusterings of high dimensional data. For datasets that are attribute rich it is often desirable to find alternative ways to group elements into cohesive clusters. For example alternative clusterings of a population of individuals according to their professional activities, shopping, and interests etc. may all yield useful groupings while a single overall clustering may be less informative when a large number of attributes are involved. Each susbpace result in RSMM provides an alternative partitioning of the data according to mixture components that define the distribution in that subspace.  Number of clusters can vary between subspaces and further analysis can be performed to cluster data according to subspace results. An example of clusterings obtained by constructing an RSMM model using the Mushroom dataset is provided in Figure~\ref{fig_component_freq}. This dataset contains the physical properties of 8124 gilled mushrooms from 23 species in the Agaricus and Lepiota family and was obtained from the UCI repository~\cite{Lichman:2013}. The model was built using one-hot encoding of data (discussed in Section \ref{mitigations}) to 118 dimensional vectors.  Using 59 random subspaces of dimensionality 4, the most common number of mixture components encountered across all subspaces was 6. In this example the number of clusters RSMM model has produced varied from 4 to 12 clusters with 6 being the most common result. It is possible to produce more refined clusterings through combination of subspace results and application of RSMM to clustering will be explored at a later work.

\subsection{Preprocessing of Attributes} \label{mitigations}
There are two additional considerations that have to be addressed to be able to build the model successfully. One of them is related to handling categorical attributes. Categorical attributes are mapped onto binary numerical attributes for each value using what is known as one-hot encoding. For example, an attribute for `Color' with values `Red', `Blue', or `Yellow' can be converted into 3 boolean attributes. This is a reversible mapping that does not cause any loss of information. While one-hot encoding leads to an increase in the dimensionality of the data, it allows for a representation that can seamlessly represent mixed numerical-categorical attributes.  

The second implementation detail is related to handling subspaces where the covariance matrices are singular. While using an ensemble of lower dimensional subspaces naturally alleviates this issue, some additional mitigations need to be added.  This is particularly the case when categoric attributes are handled using one-hot encoding. We use a numerical trick to avoid singularities by introducing noise to the data after standardizing the data.  The data is standardized by transforming each attribute using a $z$-transformation:
\begin{equation}
z= (x-\mu)/\sigma
\end{equation}
where $\mu$ and $\sigma$ respectively represent the mean and standard deviation of an attribute according to the global distribution. After this transformation we vary each coordinate by a small amount of noise:
\begin{equation}
z\rightarrow z+ \eta
\end{equation} 
where $\eta$ is distributed according to a normal distribution $N(0,\gamma)$. Because the noise is added in $z$ space, $\gamma$ is a dimensionless and universal parameter of the model. The choice of $\gamma$ does not play a significant role as long as it is small compared to 1. A value of 0.01, for example, can be used safely to avoid singularities without adding significant noise to the data.

\begin{wrapfigure}{r}{0.5\linewidth}
\vspace{-45pt}
\begin{minipage}{0.5\textwidth}
\begin{algorithm}[H]
	\caption{Random Subspace Mixture Model}
	\label{alg:Algo}
	\begin{algorithmic}[1]
		\Require $X_{s \times n}$: The training data, $k:$ Dimensionality of a subspace, $m:$ Number of desired subspaces, $\gamma:$ Noise standard deviation (default 0.01), $n\_init$: The number of initializations to perform (default 3)
		\State Set $m$ to the next closest value that satisfies Equation \ref{setm}
		\State Set $\hat{S}$ to be a set of $m$ randomly generated distinct subspaces of dimension $k$ from $n$.
		\For {i=1,2,...,$m$}
		\State \qquad Standardize and Add Noise with $\gamma$
		\State \qquad Build $gmm_{i}^{c}(x)$ using $\hat{S}[i]$ with $n\_init$ and $c=1$
		\State \qquad Calculate BIC value using $gmm_{i}^{c}(x)$ with $c=1$
		\State \qquad Increment $c$ and Build GMM until BIC increases
		\State \qquad Store best result, $gmm_{i}^{\ast}(x)$, according to BIC
		\EndFor
	\end{algorithmic}
\end{algorithm}
\end{minipage}
\end{wrapfigure}

\subsection{Implementation of the Algorithm}
One of the most appealing aspects of this algorithm is the simplicity of its implementation. The problem in each subspace can be solved using any of the existing mixture model implementations and the overall result is given by a combination of these results as defined in Equation~\ref{geometric}. The algorithm has easily been implemented in both java and python. The java implementation uses a custom Expectation Maximization algorithm initializing the clusters with k-means result from Apache commons math library and the python implementation has leveraged the GaussianMixture implementation provided by scikit-learn~\cite{scikit-learn,sklearn_api}. The python implementation also builds off of the PyOD toolkit~\cite{JMLR:v20:19-011}. A description of the steps of RSMM is provided in Algorithm~\ref{alg:Algo}.

\section{Related Work}
\label{related_work}
The focus of this paper is on the use of RSMMs for unsupervised anomaly detection. The two primary components of RSMM, namely mixture models and subspace-based methods, will be reviewed in their use for anomaly detection. We will also review related works in model interpretability.

\subsection{Mixture Models for Anomaly Detection} 

Mixture models have a long history of being used to detect anomalies~\cite{barnett1994outliers,chandola2009anomaly,pimentel2014review,riveiro2018maritime,ettabaa2019anomaly} for both numerical and categorical data~\cite{Savkli2017GALILEOAG}. Various techniques have been used to determine the number of components for a mixture model such as greedy EM-learning~\cite{laxhammar2008anomaly}, minimal message length~\cite{greggio2018anomaly}, stacking~\cite{kumar2019detecting}, holdout techniques~\cite{laxhammar2009anomaly}, and assumptions on the components distribution~\cite{kumar2019detecting}. Similar to our approach, BIC is used in~\cite{li2016anomaly} but a predefined set of $c$ are evaluated and the model with the overall minimum BIC is selected. To address singularities, assumptions on the priors have been utilized ~\cite{greggio2018anomaly} as well as thresholds on the variances~\cite{wang2019nonlinear}.  %Most methods build a single model for normal and anomalous data either assuming the data is all normal or accounting for the contamination but some create separate models for the normal data and the anomalous data~\cite{eskin2000anomaly, bahrololum2008anomaly}.  

Another use of mixture models for anomaly detection has been in conjunction with dimensionality reduction techniques. Our approach reduces the dimensionality of the models by aggregating the probability densities of an ensemble of low-dimensional subspaces which naturally handles singularities due to high-dimensionality and allows for the ability to ascribe an anomaly to specific attributes within an anomalous state vector.

\subsection{Subspace Methods for Anomaly Detection}

Subspace-based methods can be described as a type of ensemble method and are a relatively new concept explored for anomaly detection~\cite{aggarwal2013outlier,zimek2014ensembles}. The most similar method to ours is Feature Bagging~\cite{lazarevic2005feature} which also looks at random subspaces but instead of looking at low-dimensional subspaces of a certain dimension $k$, it sets a range of dimensions from $n/2$ to $n-1$ to randomly select from. Additional differences are that Feature Bagging allows duplicate subspaces to be used, it does not require equal representation of each attribute across the subspaces, the base algorithm is Local Outlier Factor (LOF)~\cite{breunig2000lof} by default, and it uses arithmetic averaging to combine scores across subspaces. Other subspace methods combine results from their subspaces using a selected average~\cite{seidl2009outlier,keller2012hics,zhao2019dcso}, harmonic mean~\cite{muller2012outlier}, weighted average~\cite{nguyen2010mining}, product~\cite{muller2011statistical}, average of normalized scores~\cite{kriegel2011interpreting}, and by using mixture models to convert to a probability density~\cite{gao2006converting}. Opposed to looking at random subspaces, other techniques used include methods that look for high-contract subspaces~\cite{keller2012hics} and subspaces that build on closely correlated dimensions~\cite{muller2012outlier,savkli2016bayesian}.

\subsection{Anomaly Interpretability} 
Model interpretability is an active area of research which has more been focused on classification and regression tasks~\cite{carvalho2019machine}.  Several methods used to explain outliers are outlined in \cite{zimek2018there}.  Additionally, \cite{sipple2020interpretable} proposed a novel method of variable attribution that uses Negative Sampling Neural Networks and Integrated Gradients to compute a proportional blame for each anomalous point and \cite{carletti2020interpretable} analyzed feature importance of the prominent Isolation Forest algorithm.  

\section{Results}

To evaluate the RSMM algorithm on its ability to detect outliers, we selected  all datasets from the Outlier Detection DataSets (ODDS) \cite{Rayana:2016} that had at least 5,000 data points and a dimesion of at least 5, summarized in Table~\ref{tab_datasets}. Additionally, we test against a real-world dataset from Smart Buildings introduced in \cite{sipple2020interpretable}. To incorporate categorical datasets, the 2 largest relevant categorical datasets from the UCI repository \cite{Dua:2019} were also incorporated.  Each dataset was constructed to have a relatively small number of anomalies, $a$, with the percentage of anomalies, $a\%=a/s$ range from $1\%$ to $32\%$ across the 12 datasets.  The Nusery dataset from the UCI repository is converted into an anomaly detection dataset by setting two smallest classes, $recommend$ and $very\_recom$, as the anomalous class.  The Mushrooms dataset from the UCI repository is converted into an anomaly detection datset by keeping all the $edible$ data points as the normal points and randomly selecting 792 of the 3,916 $poisonous$ data points to be the anomalies.  Datasets with at least 5,000 data points were selected in order to test the performance of the algorithms on their ability to perform with a small percentage of training data (down to $5\%$), which required a larger starting size.  To select the default parameters of the algorithms, 3 additional datasets from the ODDS  repository were selected based on their varying range in dimensions.

All results are generated by randomly splitting the data into 60\% training data and 40\% testing data. 10 iterations are run for each dataset. The results are presented as ROC (i.e., True Positive vs. False Positive) AUC and were obtained on a MacBook Pro with a 2.9 GHz Intel Core i7 Processor.

\begin{wraptable}{r}{0.5\linewidth}
%\begin{table}[thbp]
	\centering
\vspace{-25pt}
		\caption{Summary of Anomaly Detection Datasets Ordered by Dimensionality.}
		\label{tab_datasets}
\scalebox{0.8}{
		\begin{tabular}{c c c c c}
			\hline
			Dataset & $s$ & $n$ & Attribute Type & $a$ $(a\%)$\\
			\hline
			Mammography (MM) & 11,183 & 6 & Numerical & 260 (2.3\%) \\
			Annthyroid (AT) & 7200 & 6 & Numerical & 534 (7.4\%) \\
			Smart Buildings (SB) & 60,425 & 8 & Numerical & 1,921 (3.2\%) \\
			Shuttle (SH) & 49,097 & 9 & Numerical & 3,511 (7.2\%) \\
			Forest Cover (FC) & 286,048 & 10 & Numerical & 2,747 (1.0\%) \\
			Pen Digits (PD) & 6,870 & 16 & Numerical & 156 (2.3\%) \\
			Satellite 1 (S1) & 6,435 & 36 & Numerical & 2,036 (32\%) \\
			Satellite 2 (S2) & 5,803 & 36 & Numerical & 71 (1.2\%) \\
			Optdigits (OD) & 5216 & 64 & Numerical & 150 (3\%) \\
			Mnist (MN) & 7603 & 100 & Numerical & 700 (9.2\%) \\
			Nursery (NS) & 12,960 & 27 (8) & Categorical & 330 (2.5\%) \\
			Mushroom (MR) & 5,000 & 117 (22) & Categorical & 792 (16\%) \\
			\hline
		\end{tabular}
}
%\end{table}
\end{wraptable}

\subsection{Subspace Selection Study} \label{subspace_study}

While random subspaces are used by our algorithm, the subspace dimension $k$ was analyzed in conjunction with the number of subspaces $m$.  Selected datasets from the ODDS not used for benchmark comparisons, such as the Thyroid ($n=6$), Breast Cancer ($n=30$), and Arrhythmia ($n=274$) datasets, were used for this study based on their varying dimensionalities. Candidate subspace dimensions were set to $k\in\{2,3,4,\sqrt{n},2\sqrt{n},3\sqrt{n}\}$ and candidate number of subspaces were set to $m\in\{n,2n,3n,4n,50,100,500\}$. It was found that lower dimensional subspaces, even as small as $k=2$, obtained the best results for many datasets and that selecting $m$ as a function of $d$ was generally better than an set integer. A benefit of using $k=2$ is that the subspaces can be easily visualized which can help with interpretability.

Special considerations are made for categorical data in which one-hot encoding is used, for example the Nursery dataset and the Mushroom dataset.  If a dataset with numerical attributes is set to $k=2$, a dataset with one-hot encoded categorical attributes is set to $k=2*\overline{c}$ where $\overline{c}$ is equal to the average number of categories an attribute has.  In future work, we will explore leaving categorical attributes unexpanded as an input to the RSMM algorithm and expanding within the subspaces so that handling mixed categorical and numerical attributes is even more seamless.

\subsection{Benchmark Comparison Study}

The ability of RSMMs (with $k=2$ and $m=3n$)\footnote{Nursery $k=8$ and Mushroom $k=12$ as discussed in the previous section} to detect anomalies on the benchmark datasets is compared against ten baseline algorithms.  Specifically, the competitors are \textbf{Clustering-Based Local Outlier Factor (CBLOF)}, \textbf{Feature Bagging (FB)}, \textbf{Histogram-based Outlier Score (HBOS)}, \textbf{Isolation Forest (IForest)}, \textbf{k-Nearest Neighbor (KNN)}, \textbf{Local Outlier Factor (LOF)}, \textbf{One-Class Support Vector Machines (OCSVM)}, \textbf{Variational AutoEncoder (VAE)}, and \textbf{Gaussian Mixture Model (GMM)}.  All of the baseline algorithms except GMM can be found in  PyOD \cite{JMLR:v20:19-011}, a popular open-source Python toolbox for performing scalable outlier detection on multivariate data with a single, well-documented API.  The GMM algorithm used is the base algorithm used for RSMM, but instead of using the subspace dimension $k$, a mixture model is built using all attibutes with dimension $n$.  All of the algorithms are run with default parameters, as we want to test for cases in which data is too large to be labelled and optimized.

As shown in Table~\ref{tab_benchmark_comparison}, RSMM is competitive with all of the other algorithms including the prominent Isolation Forest algorithm.  The RSMM algorithm scores the highest in terms of ROC-AUC with a score of $85.23\%$  which is $2.3\%$ higher than IForest  at $82.87\%$.  RSMM ranks first in half of the datasets, more than any other algorithm.  It is also interesting to observe that the performance of RSMM on the benchmark datasets did not degrade with an increase in the dimensionality of the datasets. Even when a small subset of the possible subspaces are selected, the RSMM algorithm scored well on the higher dimensional datasets. RSMM also performs best on the categorical datasets.

%\newgeometry{margin=2.5cm} % modify this if you need even more space
%\begin{landscape}
	
\begin{table*}[thb]
	\begin{center}
		\caption{Mean and standard deviation of ROC-AUC scores  as \% for benchmark datasets (mean and standard deviation is calculated using 10 independent trials, highest mean highlighted in bold); Rank of mean is shown in parenthesis (lower is better). RSMM outperforms all baselines.}
		\label{tab_benchmark_comparison}
		\scalebox{0.9}{
		\begin{tabular}{c | c c c c c c c c c c | c}
			\hline
			\textbf{Data} & \textbf{CBLOF} & \textbf{FB} & \textbf{HBOS} & \textbf{IForest} & \textbf{KNN} & \textbf{LOF}  & \textbf{OCSVM} & \textbf{COPOD} & \textbf{VAE} & \textbf{GMM} & \textbf{RSMM} \\
			\hline
			
		MM&82$\pm$2 (9)&75$\pm$2 (10)&84$\pm$2 (6)&86$\pm$2 (4)&84$\pm$2 (6)&73$\pm$3 (11)&87$\pm$2 (3)&\textbf{91$\pm$2 (1)}&89$\pm$1 (2)&84$\pm$2 (6)&86$\pm$2 (4)\\
		AT&66$\pm$2 (10)&79$\pm$4 (4)&61$\pm$3 (11)&82$\pm$2 (3)&79$\pm$1 (4)&72$\pm$2 (7)&68$\pm$2 (8)&77$\pm$1 (6)&67$\pm$2 (9)&88$\pm$1 (2)&\textbf{90$\pm$1 (1)}\\
		SB&80$\pm$1 (7)&48$\pm$4 (11)&93$\pm$1 (3)&81$\pm$4 (6)&70$\pm$1 (9)&55$\pm$2 (10)&85$\pm$0 (5)&\textbf{94$\pm$0 (1)}&90$\pm$0 (4)&80$\pm$9 (7)&\textbf{94$\pm$1 (1)}\\
		SH&61$\pm$2 (9)&48$\pm$3 (11)&99$\pm$0 (2)&\textbf{100$\pm$0 (1)}&65$\pm$0 (8)&53$\pm$1 (10)&99$\pm$0 (2)&99$\pm$0 (2)&99$\pm$0 (2)&84$\pm$9 (7)&98$\pm$0 (6)\\
		FC&94$\pm$0 (2)&57$\pm$2 (10)&71$\pm$0 (9)&88$\pm$2 (5)&78$\pm$1 (8)&54$\pm$1 (11)&\textbf{95$\pm$0 (1)}&88$\pm$0 (5)&93$\pm$0 (3)&90$\pm$1 (4)&87$\pm$1 (7)\\
		PD&88$\pm$4 (7)&47$\pm$4 (10)&92$\pm$1 (5)&\textbf{95$\pm$1 (1)}&75$\pm$2 (9)&47$\pm$4 (10)&93$\pm$1 (3)&91$\pm$1 (6)&94$\pm$1 (2)&76$\pm$6 (8)&93$\pm$1 (3)\\
		S1&75$\pm$1 (3)&56$\pm$2 (10)&76$\pm$1 (2)&71$\pm$1 (5)&68$\pm$1 (6)&56$\pm$2 (10)&66$\pm$1 (8)&63$\pm$1 (9)&68$\pm$5 (6)&73$\pm$2 (4)&\textbf{80$\pm$1 (1)}\\
		S2&\textbf{100$\pm$0 (1)}&46$\pm$10 (10)&98$\pm$1 (6)&\textbf{100$\pm$0 (1)}&95$\pm$1 (8)&46$\pm$10 (10)&\textbf{100$\pm$0 (1)}&98$\pm$1 (6)&99$\pm$1 (5)&94$\pm$1 (9)&\textbf{100$\pm$0 (1)}\\
		OD&77$\pm$5 (3)&45$\pm$6 (9)&\textbf{87$\pm$1 (1)}&71$\pm$5 (4)&37$\pm$2 (11)&45$\pm$6 (9)&50$\pm$2 (8)&68$\pm$1 (5)&51$\pm$1 (7)&55$\pm$5 (6)&86$\pm$1 (2)\\
		MN&\textbf{85$\pm$1 (1)}&72$\pm$2 (8)&57$\pm$1 (11)&80$\pm$3 (6)&\textbf{85$\pm$1 (1)}&72$\pm$2 (8)&\textbf{85$\pm$1 (1)}&77$\pm$1 (7)&\textbf{85$\pm$1 (1)}&81$\pm$6 (5)&58$\pm$2 (10)\\
		NS&50$\pm$8 (6)&55$\pm$5 (2)&49$\pm$20 (8)&51$\pm$8 (3)&49$\pm$8 (8)&51$\pm$4 (3)&49$\pm$3 (8)&46$\pm$8 (11)&50$\pm$19 (6)&51$\pm$6 (3)&\textbf{57$\pm$9 (1)}\\
		MR&84$\pm$2 (8)&54$\pm$4 (10)&87$\pm$1 (5)&91$\pm$1 (3)&92$\pm$1 (2)&52$\pm$3 (11)&86$\pm$1 (7)&88$\pm$1 (4)&87$\pm$1 (5)&74$\pm$18 (9)&\textbf{94$\pm$1 (1)}\\
		\hline
		\textbf{AVG}&79$\pm$2 (7)&57$\pm$4 (10)&80$\pm$3 (6)&83$\pm$3 (2)&73$\pm$2 (9)&56$\pm$3 (11)&80$\pm$1 (5)&82$\pm$1 (3)&81$\pm$3 (4)&78$\pm$6 (8)&\textbf{85$\pm$2 (1)}\\
			
			\hline
		\end{tabular}
		}
	\end{center}
\vspace{-25pt}
\end{table*}

\subsection{Small Percent Training Study}

Datasets with at least 5,000 data points were chosen so that a study could be conducted to determine how well the algorithms performed with only a small fraction of the data used for training.  It was found that most all the algorithms performed just as well with down to $5\%$ of the data used for training (compared to the $60\%$ used in the previous study).  Most algorithms did not perform statistically significantly differently and the ranking of the top 5 algorithms did not change.

%\end{landscape}
%\restoregeometry

\subsection{Anomaly Interpretability Study}

The analysis on the interpretability of the RSMM algorithm for anomaly detection will be focused on the Smart Buildings dataset which was chosen because the attribute names were easily accessible. A similar analysis could be done on any dataset. 

\begin{figure*}[thb]
\vspace{-15pt}
	\centerline{\includegraphics[width=.8\textwidth]{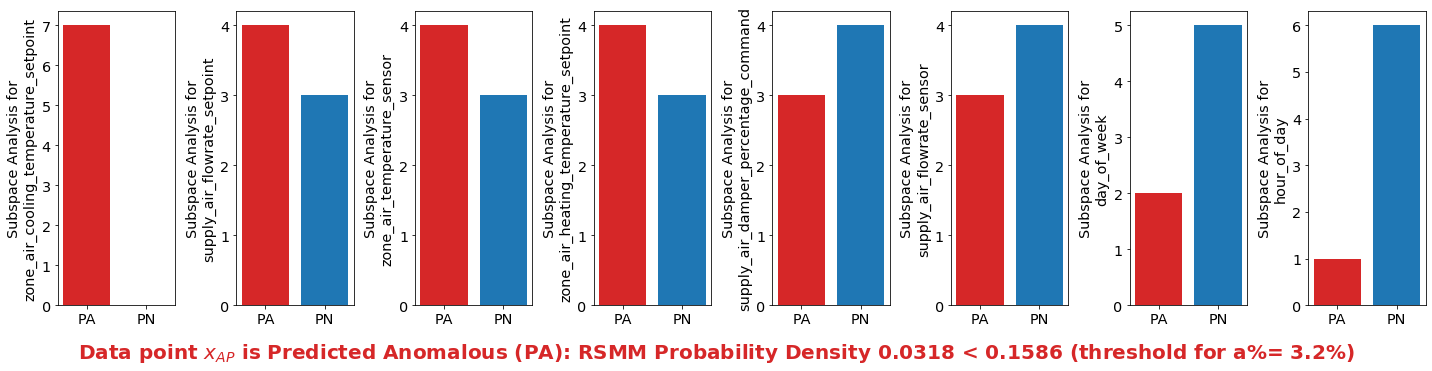}}
	\caption{Anomaly Interpretation of an \emph{Anomalous} point $x_{AP}$. Each bar plot represents an analysis of the subspaces for a specific attribute, tallying the number of subspaces the attribute is predicted anomalous (PA) compared to the number of subspaces the attribute is predicted normal (PN).  The attributes that are most often predicted anomalous in the subspaces are sorted towards the left. `zone\_air\_cooling\_temperature\_setpoint' is predicted anomalous for every subspace it is apart of for $x_{AP}$.}
	\label{fig_interpretation_anom}
\end{figure*}

\begin{figure}[thb]
\vspace{-45pt}
	\centerline{\includegraphics[width=.7\columnwidth]{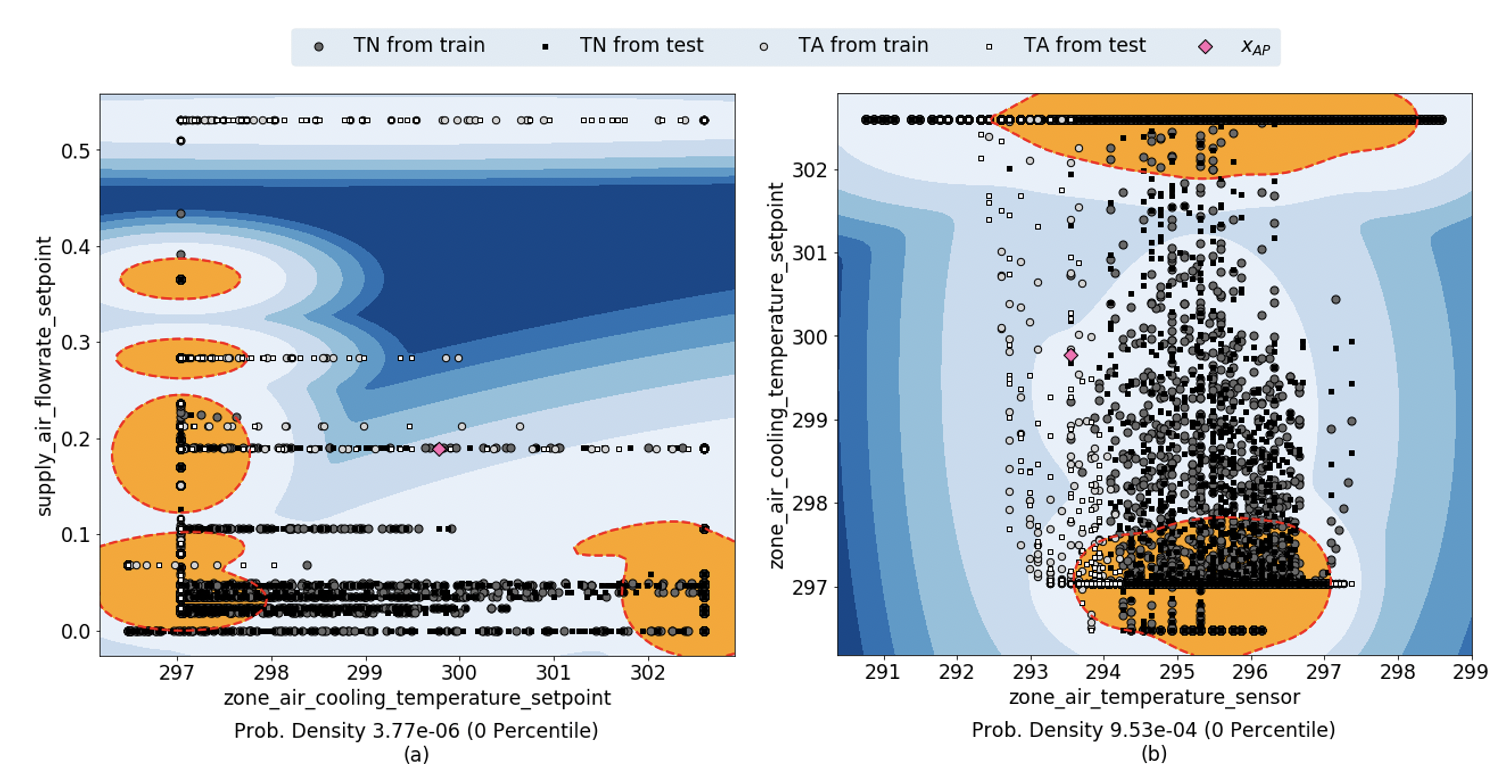}}
	\caption{View of the \emph{Anomalous} point $x_{AP}$ from Figure~\ref{fig_interpretation_anom} in the two subspaces with the lowest probability density percentiles for $x_{AP}$.  The shaded contours illustrate the different regions of probability densities for the subspaces, with dashed lines drawn at the decision threshold for an anomaly rate of 3.2\% (the a\% of the Smart Buildings dataset).}
	\label{fig_subspaces_anom}
\end{figure}

While obtaining a high AUC is valuable for an anomaly detection algorithm, understanding \textit{why} a data point is anomalous is extremely important in determining how to handle it. A benefit of RSMM is that analyses can be conducted across the subspaces to understand how much each attribute contributes to a given data point being anomalous. While each subspace has different ranges in probability densities, for example, the range in probability densities of the training data for the subspace in Figure~\ref{fig_subspaces_anom}(a) is $(4.69\textrm{e-}13, 3.39\textrm{e-}0)$ whereas the range in probability densities of the training data for the subspace in Figure~\ref{fig_subspaces_anom}(b) is $(1.74\textrm{e-}4, 9.06\textrm{e-}1)$, a decision threshold can be determined for each subspace based on the anomaly rate of the dataset a\%. For a given data point, the attributes can be ranked based on the number of subspaces it is a part of that predicts the point to be anomalous (PA) compared to the number of subspaces it is a part of that predicts the point to be normal (PN). The comparison of PA to PN across attributes is fair since all subspaces are generated randomly and have the same dimension $k$.

Figure~\ref{fig_interpretation_anom} ranks the 8 attributes in the Smart Buildings dataset for an anomalous point $x_{AP}$ as decribed in the previous paragraph, with the attributes that are most often predicted anomalous in the subspaces sorted towards the left. Another way to view why this data point is anomalous is by looking at the most anomalous subspaces, as seen in Figure~\ref{fig_subspaces_anom}, where (a) is the subspace with the lowest probability density percentile compared to the training data for this point and (b) is the second lowest. All True Normal (TN) and True Anomalous (TA) points from the train and test data are plotted.  The shaded contours illustrate the different regions of probability densities for the subspaces, with dashed lines drawn at the decision threshold for the a\% of the dataset. $x_{AP}$ is clearly an outlier in both of these subspaces.

\section{Future Work}

RSMM is naturally suited to be used for classification purposes. Applications of probabilistic models to classification requires a specification of class likelihood ratios. It is possible to follow an analogous statistical approach to evaluate the likelihood ratios of classes by defining them in terms of geometric averaging of subspace likelihood ratios. Preliminary tests of classification have provided highly accurate results on some well-known datasets and a detailed study of classification will be considered in a future work. Additionally, an important potential application involves clustering of high dimensional data. Gaussian mixture models naturally provide the structure of clusters in the data. In the case of RSMM, each subspace result provides an alternative clustering of the data which is a desirable perspective to understand complex relationships between the data elements and it is a direction that will be explored in a future work. 

\section{Conclusions}

In this paper, we have presented RSMM, a new method of generating probabilistic models for attribute rich datasets with numerical, categorical, or mixed attributes. The RSMM is based on constructing a global distribution using geometric averaging of subspace probability densities in which attributes are equally represented across the subspaces. Each subspace distribution is constructed with a Gaussian mixture model using an optimality criterion such as BIC/AIC. RSMM is trivially parallelizable and simple to implement yet performs as well as some of the best existing algorithms. Beyond its computational advantages and its accuracy, RSMM also provides an interpretable result that gives insights about the causes of anomalies.

\section{Acknowledgments}
This work was supported by internal research and development funding provided by JHU/APL.

%{
%\fontsize{9pt}{10pt}
%\selectfont

\end{document}